\documentclass[pdflatex,sn-mathphys-num]{sn-jnl}% Math and Physical Sciences Numbered Reference Style
\usepackage{graphicx}%
\usepackage{multirow}%
\usepackage{amsmath,amssymb,amsfonts}%
\usepackage{amsthm}%
\usepackage{mathrsfs}%
\usepackage[title]{appendix}%
\usepackage{xcolor}%
\usepackage{textcomp}%
\usepackage{manyfoot}%
\usepackage{booktabs}%
\usepackage{algorithm}%
\usepackage{algorithmicx}%
\usepackage{algpseudocode}%
\usepackage{listings}%
\usepackage{svg}%
\usepackage{xcolor}
\usepackage{xcolor}
\usepackage{soul}

\sethlcolor{green!20}
\theoremstyle{thmstyleone}%
\theoremstyle{thmstyletwo}%

\theoremstyle{thmstylethree}%

\begin{document}

\title[Lesion-Guided ROI Deep Learning]{Less Contouring, More Accuracy: Lesion-Guided ROI Deep Learning for Ovarian Ultrasound Classification}

\author[1,2,3]{\fnm{Mehran} \sur{Ahmad}}

\author[1]{\fnm{Ali} \sur{Abbasian Ardakani}}

\author[4]{\fnm{Afshin} \sur{Mohammadi}}  

\author[1]{\fnm{Alisa} \sur{Mohebbi}}

\author[2]{\fnm{Gernot} \sur{Kronreif}}

\author*[1,2,3]{\fnm{Sepideh} \sur{Hatamikia}}
\email{sepideh.hatamikia@dp-uni.ac.at}

\affil[1]{\orgdiv{Department of Medicine},
\orgname{Danube Private University (DPU)},
\orgaddress{\city{Krems}, \country{Austria}}}

\affil[2]{\orgname{Austrian Centre for Medical Innovation and Technology (ACMIT)},
\orgaddress{\city{Wiener Neustadt}, \country{Austria}}}

\affil[3]{\orgdiv{Department of Medical Physics and Biomedical Engineering},
\orgname{Medical University of Vienna},
\orgaddress{\city{Vienna}, \country{Austria}}}

\affil[4]{\orgdiv{Department of Radiology, Faculty of Medicine},
\orgname{Urmia University of Medical Science},
\orgaddress{\city{Urmia}, \country{Iran}}}


\abstract{The abstract serves both as a general introduction to the topic and as a brief, non-technical summary of the main results and their implications. Authors are advised to check the author instructions for the journal they are submitting to for word limits and if structural elements like subheadings, citations, or equations are permitted.}

%%================================%%
%% Sample for structured abstract %%
%%================================%%

\abstract{
\textbf{Objective:} The classification and malignancy risk assessment of ovarian lesions using transvaginal ultrasound are challenging because of overlapping imaging features and operator-dependent interpretation. This study compared deep learning (DL) and machine learning (ML)-based radiomics strategies to identify an approach that maintains high diagnostic performance while minimizing annotation burden.

\textbf{Methods:} Two retrospective ovarian ultrasound datasets were used: the Multi-Modality Ovarian Tumor Ultrasound (MMOTU) dataset for eight-class classification and the Ovarian Ultrasound Dataset (OUD) for binary classification. Four strategies were evaluated under a unified experimental protocol: global image-based DL, lesion-guided region-of-interest (ROI)-based DL, lesion contour-based DL, and lesion contour-based radiomics with conventional ML classifiers. Four DL architectures, MaxViT-Tiny, Swin Transformer, EfficientNet-B7, and ResNet18, were evaluated.For the radiomics strategy, quantitative features were extracted from mask-isolated lesion regions and classified using a radial basis function support vector machine (SVM), k-nearest neighbors (KNN), and an artificial neural network (ANN). For the lower-sample OUD dataset, ANOVA-based feature selection was applied to reduce the initial 215-dimensional feature space before classification. Bootstrap resampling was used for robust performance evaluation.

\textbf{Results:} The analysis included 507 patient cases across the MMOTU and OUD datasets. The lesion-guided ROI-based DL strategy achieved the strongest overall performance. MaxViT-Tiny achieved 93.10\% accuracy with an AUC of 0.99 on MMOTU and 97.56\% accuracy with an AUC of 0.99 on OUD. The lesion contour-based approach achieved comparable performance, with 91.81\% accuracy and an AUC of 0.99 on MMOTU and 95.12\% accuracy and an AUC of 0.99 on OUD, but required substantially greater annotation effort because of pixel-level contour delineation. Radiomics-based models showed competitive performance in the binary classification setting but were less effective for multiclass classification.

\textbf{Conclusion:} Lesion-guided ROI-based DL provided the best balance between diagnostic performance and annotation burden for ovarian ultrasound classification. This strategy may offer a more practical and scalable approach for AI-assisted ovarian lesion assessment and clinical workflow integration. Prospective multicentre validation is required to confirm its broader generalizability and clinical utility.
}




\maketitle

\section{Introduction}\label{sec1}

Transvaginal sonography (TVS) serves as the primary imaging modality for the evaluation of ovarian masses, providing real-time, high-resolution visualization of lesion morphology and internal composition \cite{bib1,bib2,bib3}. In clinical practice, TVS enables risk stratification through standardized systems such as the International Ovarian Tumor Analysis (IOTA) models and the Ovarian-Adnexal Reporting and Data System (O-RADS), thereby supporting clinical management decisions \cite{bib4,bib5}. However, ultrasound interpretation remains highly operator-dependent, with substantial inter-observer variability arising from overlapping sonographic features between benign and malignant lesions and subtle distinctions among specific ovarian lesion etiologies \cite{bib6,bib7}. These limitations highlight the need for objective artificial intelligence (AI)-based adjunctive tools to improve diagnostic consistency and reduce subjectivity in radiological workflows.

Both DL and radiomics-based models applied to TVS have demonstrated considerable potential for assisting radiologists in ovarian lesion characterization and etiological classification. DL architectures can be integrated into clinical reporting workflows to provide rapid and objective probability estimates for malignancy and for specific diagnoses, such as serous cystadenoma, teratoma, or high-grade serous carcinoma. Such tools may help reduce inter-observer variability and support less experienced examiners in lesion triage, follow-up planning, and timely referral \cite{bib8,bib9,bib10,bib11,bib12,bib13}.

Radiomics-based approaches complement DL by providing quantitative and potentially interpretable imaging biomarkers derived from lesion texture, shape, and intensity characteristics. These features are typically extracted from segmented lesion regions and may be incorporated into existing O-RADS or IOTA-based workflows to refine risk stratification when visual characteristics overlap and to support etiological classification without requiring highly complex computational architectures \cite{bib8,bib14,bib15,bib16}.

Recent studies have further demonstrated the potential of AI for ovarian ultrasound diagnosis in both binary and multiclass classification settings. For binary diagnosis, a deep convolutional neural network (DCNN) trained on pelvic ultrasound data achieved an area under the receiver operating characteristic curve (AUC) of 0.911 \cite{bib17}. Feature-fusion and decision-fusion strategies applied to multimodal ultrasound achieved an AUC of 0.93 with a sensitivity of 92\% \cite{bib18}. An ensemble of ten pretrained convolutional neural networks (CNNs) achieved an accuracy of 92.15\% and specificity of 92.92\%, outperforming individual models \cite{bib19}. A CNN combined with a denoising convolutional autoencoder, achieved 97.2\% accuracy for normal-versus-tumor classification and 90.12\% accuracy for malignancy discrimination \cite{bib20}. DenseNet achieved a test accuracy of 97.5\% for benign-versus-malignant classification \cite{bib21}, while a Swin Transformer demonstrated expert-level performance on TVS, achieving an AUC of 0.92 \cite{bib22}.

The field has subsequently progressed toward multiclass ovarian lesion classification, which is more challenging but potentially more clinically informative because it aims to distinguish specific ovarian conditions rather than only separate benign from malignant lesions. The MMOTU benchmark was introduced with baseline results in which EfficientNetV2-M achieved a top-1 accuracy of 80.60\% \cite{bib23}. EfficientOvaNet, a dual-branch EfficientNet-B3 framework combining lesion ROI and global image features, achieved an accuracy of 91.9\% \cite{bib24}. However, performance across multiclass studies remains variable. A fine-tuned YOLOv8x reached only 70.6\% mAP50 for eight-class classification \cite{bib25}, while ResNet18 achieved an accuracy of 76.2\% and an F1-score of 78.2\% on the three-class (OUD) dataset, comprising normal ovary, dominant follicle (DF), and polycystic ovary (PCO) images \cite{bib26}. Collectively, these studies indicate that DL can achieve high performance in binary diagnosis, whereas multiclass recognition remains more sensitive to dataset size, class composition, lesion heterogeneity, and model design.

\begin{figure}[htbp]
\centering
\includegraphics[width=\linewidth]{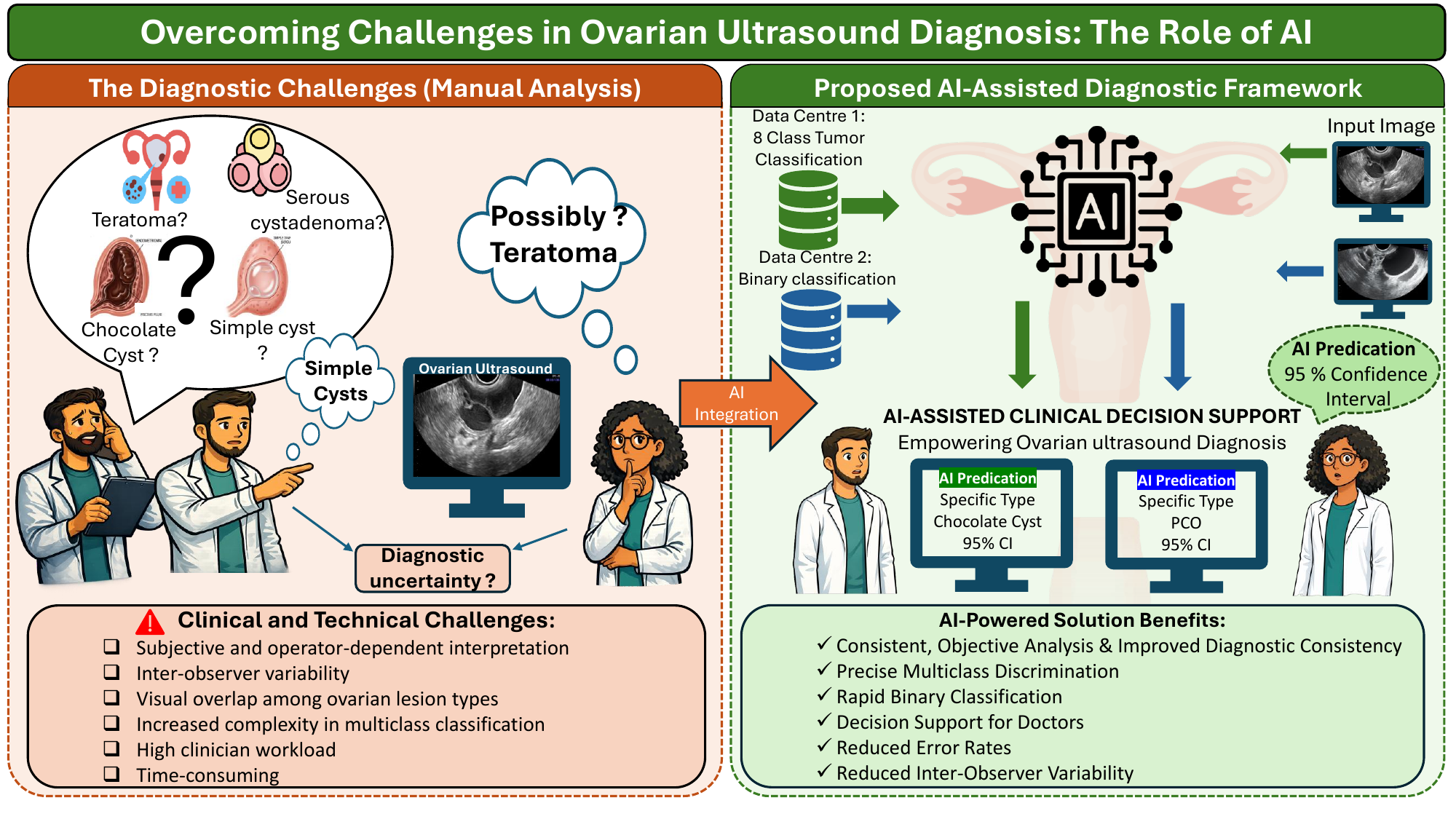}
\caption{Clinical challenges in visual ovarian ultrasound interpretation and the role of AI-assisted diagnosis.}
\label{fig1}
\end{figure}

Despite these promising findings, several important gaps remain that directly affect the practical development and implementation of AI-assisted ovarian ultrasound analysis. Most existing studies have not systematically compared global image analysis with clinically practical lesion-focused annotation strategies, such as a rapidly defined rectangular ROI around the lesion versus detailed pixel-level contour delineation. Furthermore, previous studies \cite{bib18,bib19,bib20,bib21,bib22,bib23,bib24,bib25,bib26} remain limited to a single-center dataset, and the assessment of a unified framework on different data from different centers and across different ovarian lesion types has not been explored. Moreover, in a busy center, the extra time needed for detailed contouring limits how easily these AI tools can be developed and adopted into routine practice. In routine clinical practice, ovarian lesion diagnosis relies predominantly on visual interpretation of ultrasound images, as illustrated in Fig.~\ref{fig1}, which may be subjective, operator-dependent, and time-consuming.

Therefore, the aim of this study was to compare four ultrasound-based AI strategies for ovarian lesion classification across binary and multiclass settings using two publicly available datasets. We performed a head-to-head evaluation of global image-based DL, lesion-guided ROI-based DL, lesion contour-based DL, and lesion contour-based radiomics combined with conventional ML classifiers. In particular, we investigated whether the less annotation-intensive lesion-guided ROI strategy could achieve diagnostic performance comparable to detailed contour-based approaches while substantially reducing annotation burden. Robust statistical evaluation with confidence intervals was additionally performed to assess the reliability of the observed performance differences and to identify a practical strategy for both malignancy-related discrimination and classification of specific ovarian conditions.

\section{Materials and methods}\label{sec2}

This study presents a comparative framework for ovarian ultrasound classification using two publicly available datasets with different levels of classification complexity. The MMOTU dataset \cite{bib23} was evaluated as an eight-class classification problem. The OUD \cite{bib26} was originally introduced as a three-class dataset comprising normal ovary, DF, and PCO images. In the present study,  due to low amount of data in the normal class, we used only DF versus PCO classification task. Both source datasets are de-identified and open access; therefore, additional ethics committee approval was not required.

\begin{figure}[htbp]
\centering
\includegraphics[width=\textwidth]{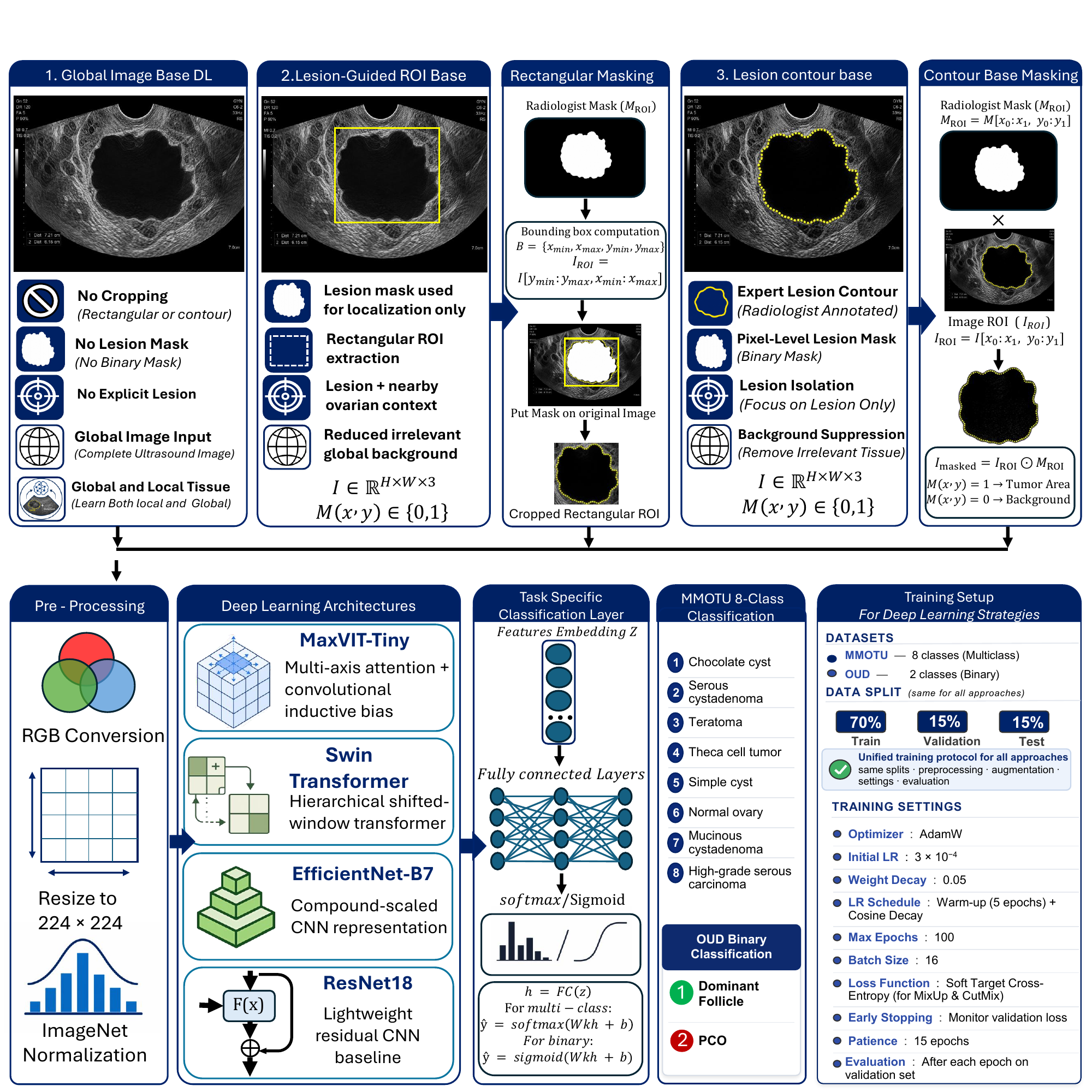}
\caption{The three deep learning strategies are based on the global ultrasound image, lesion-guided ROI, and expert-annotated lesion contour.}
\label{fig2}
\end{figure}

\begin{figure}[htbp]
\centering
\includegraphics[width=\textwidth]{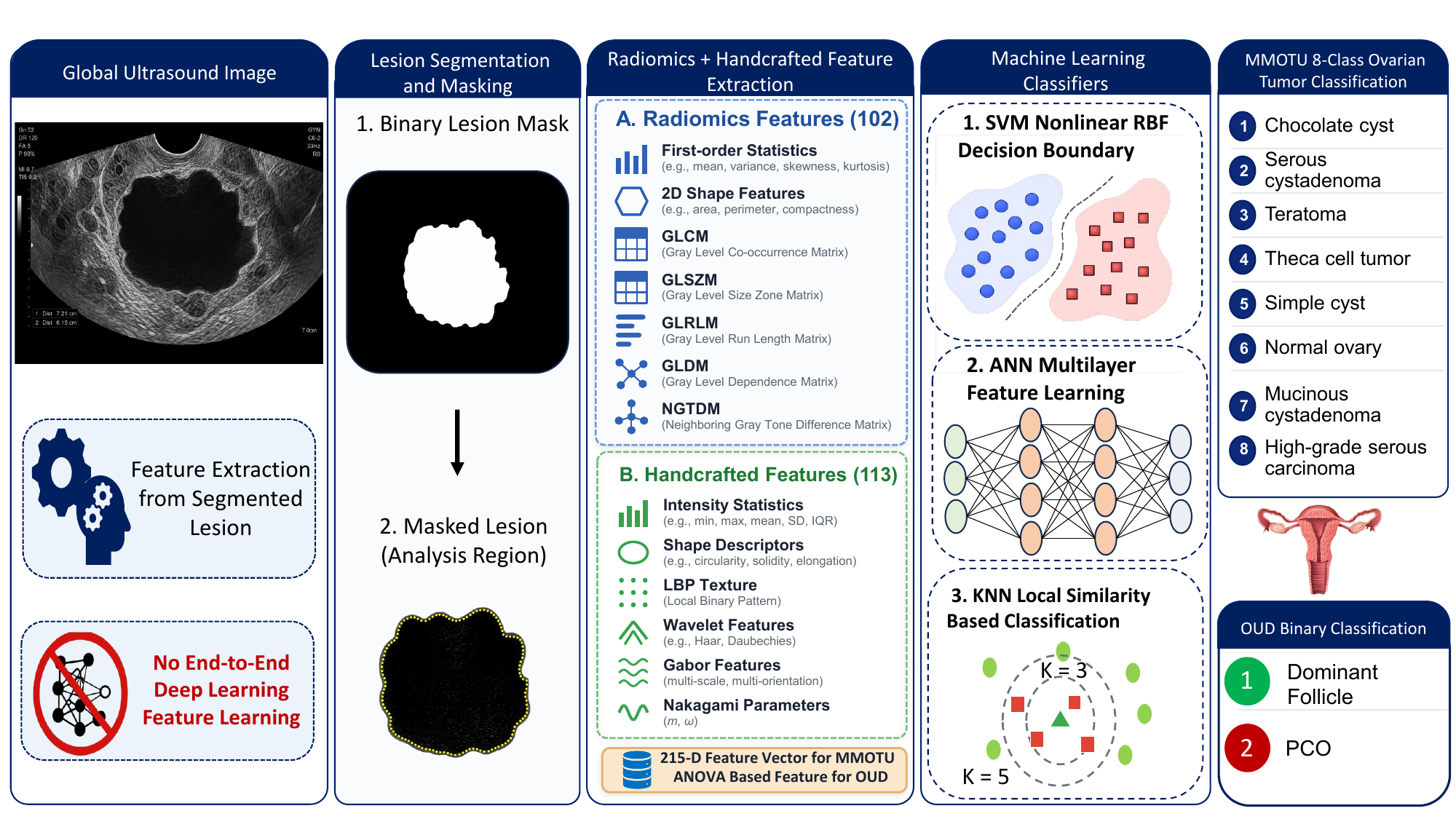}
\caption{The lesion contour-based radiomics strategy, including lesion masking, extraction of PyRadiomics and handcrafted ultrasound features, feature integration, and classification using SVM, KNN, and ANN.}
\label{fig3}
\end{figure}

Four classification strategies were investigated, as illustrated in Fig.~\ref{fig2},~\ref{fig3}: (A) global image-based DL, using the entire ultrasound image; (B) lesion-guided ROI-based DL, using a cropped region surrounding the lesion; (C) lesion contour-based DL, using only the radiologist-annotated lesion region, as illustrated in Fig.~\ref{fig2}; and (D) lesion contour-based radiomics, in which quantitative features were extracted from the segmented lesion region, as illustrated in Fig.~\ref{fig3}. The first three strategies were evaluated using four DL architectures: MaxViT-Tiny, EfficientNet-B7, Swin Transformer, and ResNet-18. For the fourth strategy, radiomics and handcrafted imaging features were classified using three ML models (SVM, KNN, and ANN). All strategies used identical dataset splits, training protocols, and evaluation metrics to ensure a fair comparison.

\subsection{Dataset description and partitioning}\label{subsec2.1}

The MMOTU dataset \cite{bib23} contains 1,469 2D ultrasound images acquired at Beijing Shijitan Hospital, Capital Medical University, from 247 patients using a Mindray Resona 8 ultrasound system. A total of 27 expert radiologists provided pixel-wise lesion masks and eight-class diagnostic labels based on pathological reports. The original MMOTU partition comprised 1,000 training images from 171 patients and 469 held-out images from 76 patients, with patient-level independence maintained between these two partitions. Because image-specific patient identifiers were not available within the 469-image held-out partition, this subset was further divided at the image level into validation ($n=237$) and final test ($n=232$) subsets. Therefore, patient-level independence between the validation and final test subsets could not be verified. Detailed class distributions are provided in Supplementary File~S1.

The OUD \cite{bib26} originally consists of 301 ultrasound examinations, with each examination corresponding to a unique patient, acquired at Fatemieh Hospital, Hamedan University of Medical Sciences, during 2023--2024. All ultrasound examinations were performed using a Philips Affiniti 50 ultrasound system by a single board-certified obstetrician-gynecologist with fellowship training in infertility. For the radiomics and lesion contour-based analyses, segmentation masks were manually generated for the DF and PCO classes. The normal-ovary class was excluded because the number of normal cases was limited. Consequently, 260 images from 260 patients were included in the binary DF-versus-PCO classification task. Detailed class distributions are provided in Supplementary File~S1.

\subsection{Proposed ovarian ultrasound classification strategies}\label{subsec2.2}

Image preprocessing and training-time augmentation procedures applied across the investigated models are described in Supplementary File~S2. Four classification strategies were implemented under a unified experimental protocol.

\textbf{(A) Global image-based DL.}
In the first strategy, classification was performed using the complete ultrasound image without lesion masking or cropping. This strategy was designed to determine whether the DL models could learn discriminative ovarian lesion patterns directly from the full ultrasound frame. Both the lesion and adjacent tissues were retained, allowing the network to capture lesion-specific characteristics together with potentially informative surrounding anatomical context.

\textbf{(B) Lesion-guided ROI-based DL.}
In the second strategy, classification was performed using a lesion-guided ROI-based DL framework that focused the learning process on the lesion region rather than on the complete ultrasound image. A rectangular ROI surrounding the lesion was extracted while preserving the lesion and its immediate local context. The objective was to reduce the influence of irrelevant background structures while retaining clinically relevant contextual information surrounding the lesion, including the cyst wall, septa, papillary projections, and solid components commonly considered during ovarian lesion assessment using IOTA and O-RADS descriptors. This strategy was specifically designed to investigate whether lesion-guided localization could improve classification performance while reducing the annotation time and expertise required for precise pixel-level lesion contouring.

\textbf{(C) Lesion contour-based DL.}
In the third strategy, ovarian ultrasound classification was performed using a lesion contour-based DL framework that restricted the model input more specifically to the expert-annotated lesion region. Pixels outside the radiologist-annotated lesion contour were excluded, thereby reducing the contribution of surrounding tissues and background structures. This strategy was designed to encourage the models to learn predominantly from lesion-specific morphological and textural characteristics. It was evaluated to determine whether precise lesion isolation could provide additional classification benefits compared with the lesion-guided ROI strategy, which retained both the lesion and its immediate surrounding context.

\textbf{(D) Lesion contour-based radiomics.}
In the fourth strategy, quantitative imaging features were extracted exclusively from the mask-isolated lesion region and subsequently classified using conventional machine learning algorithms. A total of 102 original PyRadiomics descriptors and 113 handcrafted ultrasound-specific features were initially extracted, resulting in a 215-dimensional feature vector. For MMOTU, all extracted features were retained for classification. For OUD, due to the limited sample size relative to the feature dimensionality, ANOVA-based feature selection was applied to rank features according to their discriminatory ability between DF and PCO classes.

 Further methodological details regarding the DL models are provided in Supplementary File~S3, while details of the ML models and the four classification strategies are provided in Supplementary File~S4. Fig.~\ref{fig2} illustrates the schematic workflow for the three DL-based strategies, whereas Fig.~\ref{fig3} presents the schematic workflow for the lesion contour-based radiomics strategy.

\subsection{Experimental protocol and model optimization}\label{subsec2.3}

To ensure a fair comparison, all classification strategies were evaluated using consistent experimental settings within each dataset. Each dataset was split into training, validation, and test sets in a 70:15:15 ratio, and the same split was used across all strategies  For the DL-based experiments, MaxViT-Tiny, Swin Transformer, EfficientNet-B7, and ResNet-18 were trained using the same optimization settings.

ResNet-18 was additionally included as an architecture-matched baseline to facilitate comparison with the original OUD study \cite{bib26}, in which ResNet-18 was reported as the best-performing architecture on the original three-class classification task comprising normal ovary, DF, and PCO images. In the present study, ResNet-18 was independently retrained under the same experimental protocol used for the reformulated binary DF-versus-PCO subset.

Model optimization was performed using the AdamW optimizer with an initial learning rate of $3 \times 10^{-4}$ and a weight decay of 0.05. A linear warm-up followed by cosine learning-rate decay was applied, with the first five epochs used for warm-up and the remaining epochs following the cosine decay schedule. All DL models were trained for a maximum of 100 epochs with a batch size of 16. Early stopping was applied based on validation loss with a patience of 15 epochs to reduce overfitting. Soft-target cross-entropy loss was used in conjunction with MixUp and CutMix augmentation. For each experiment, the checkpoint achieving the highest validation accuracy was retained for final evaluation on the corresponding test set.

For the radiomics and handcrafted feature-based framework, multiple machine learning classifiers were evaluated, including SVM, ANN, and KNN, as discussed in Supplementary File~S4.

All experiments were implemented in Python 3.11.9 using PyTorch 2.6.0, torchvision 0.21.0, CUDA 12.4, timm 1.0.25, NumPy 2.4.3, Pillow 12.1.1, scikit-learn 1.8.0, and Matplotlib 3.10.8. Experiments were conducted on Windows 11 using an NVIDIA RTX 4000 Ada Generation GPU.

\subsection{Performance evaluation and statistical analysis}\label{subsec2.4}

The performance of the proposed methods was evaluated on the corresponding test sets using top-1 accuracy, top-2 accuracy, specificity, sensitivity (recall), and AUC. For OUD radiomics, one-way ANOVA was used to rank the 215 candidate features by their ability to discriminate between DF and PCO classes. Feature selection and preprocessing were fitted exclusively on the training partition and then fixed for validation and test evaluation. Top-1 accuracy indicated model’s highest-probability prediction, whereas top-2 accuracy indicated whether the correct class was included among the two most probable predictions. Moreover, for each class, recall measured how many true samples of that class were correctly identified, and specificity measured how many samples from the remaining classes were correctly excluded. For the multiclass MMOTU dataset, AUC was computed using a one-vs-rest strategy, in which each class was compared against all remaining classes, with higher values indicating better performance.

The principal evaluation metrics were calculated as follows:

\begin{equation}
\mathrm{Accuracy} =
\frac{TP + TN}{TP + TN + FP + FN}
\label{eq:accuracy}
\end{equation}

\begin{equation}
\mathrm{Sensitivity} =
\frac{TP}{TP + FN}
\label{eq:sensitivity}
\end{equation}

\begin{equation}
\mathrm{Specificity} =
\frac{TN}{TN + FP}
\label{eq:specificity}
\end{equation}

where $TP$, $TN$, $FP$, and $FN$ denote true positives, true negatives, false positives, and false negatives, respectively. 

To quantify performance uncertainty, bootstrap resampling with 1,000 iterations was applied to the test-set predictions. In each iteration, samples were randomly drawn with replacement from the corresponding test set, and the evaluation metrics were recalculated. The resulting bootstrap distributions were used to estimate 95\% confidence intervals using the percentile method.

\section{Results}\label{sec3}

The final analysis included 1,729 ultrasound images from 507 patients across the MMOTU and OUD datasets. MMOTU contributed 1,469 images from 247 patients, whereas OUD contributed 260 images from 260 patients. Detailed class-wise distributions across the training, validation, and test sets are provided in Supplementary File~S1.

\subsection{Global image-based deep learning performance}\label{subsec3.1}

When the complete ultrasound image was used without explicit lesion localization, performance differed according to dataset complexity as shown in Table~\ref{tab1}. On the eight-class MMOTU dataset, top-1 accuracy across the evaluated architectures ranged from 71.1\% to 81.6\%, top-2 accuracy from 85.8\% to 91.0\%, sensitivity from 71.1\% to 81.5\%, and AUC from 0.93 to 0.95. MaxViT-Tiny achieved the highest top-1 accuracy of 81.6\%.

On the binary OUD dataset, top-1 accuracy ranged from 85.4\% to 95.1\%, sensitivity from 85.4\% to 95.1\%, and AUC from 0.87 to 0.99. MaxViT-Tiny again achieved the highest top-1 accuracy, reaching 95.1\%. However, the confidence intervals across architectures showed considerable overlap on the smaller OUD test set, limiting firm conclusions regarding differences among individual models. Overall, global context proved informative but insufficient: the moderate multiclass accuracy suggests that background tissue and acquisition artifacts limit the separation of overlapping tumor morphologies, restricting reliability for detailed etiological assessment in routine practice.

 \begin{table}[htbp]
\caption{Performance of the global image-based deep learning models on the MMOTU and OUD datasets. Values in brackets represent 95\% confidence intervals.}
\label{tab1}%

\centering
\fontsize{6.5}{7.5}\selectfont
\setlength{\tabcolsep}{2pt}
\renewcommand{\arraystretch}{1.15}

\begin{tabular}{@{}llccccc@{}}
\toprule

\textbf{Data} &
\textbf{Model} &
\shortstack{\textbf{Top-1}\\\textbf{Accuracy (\%)}} &
\shortstack{\textbf{Top-2}\\\textbf{Accuracy (\%)}} &
\shortstack{\textbf{Specificity}\\\textbf{(\%)}} &
\shortstack{\textbf{Sensitivity}\\\textbf{(\%)}} &
\textbf{AUC} \\

\midrule

\textbf{MMOTU}
& MaxViT-Tiny
& \shortstack{81.57\\{[76.72, 85.78]}}
& \shortstack{90.09\\{[86.21, 93.53]}}
& \shortstack{96.69\\{[90.80, 100.00]}}
& \shortstack{81.57\\{[76.72, 85.78]}}
& \shortstack{0.94\\{[0.92, 0.96]}} \\

& Swin
& \shortstack{78.45\\{[73.28, 83.62]}}
& \shortstack{91.00\\{[86.21, 93.55]}}
& \shortstack{96.22\\{[89.23, 100.00]}}
& \shortstack{78.45\\{[73.28, 83.62]}}
& \shortstack{0.95\\{[0.9386, 0.97]}} \\

& EfficientNet-B7
& \shortstack{79.31\\{[74.14, 84.48]}}
& \shortstack{89.66\\{[85.34, 93.10]}}
& \shortstack{96.24\\{[89.48, 100.00]}}
& \shortstack{79.31\\{[74.14, 84.48]}}
& \shortstack{0.95\\{[0.93, 0.97]}} \\

& ResNet18
& \shortstack{71.12\\{[65.09, 76.72]}}
& \shortstack{85.78\\{[80.60, 90.09]}}
& \shortstack{95.68\\{[88.43, 99.01]}}
& \shortstack{71.12\\{[65.09, 76.72]}}
& \shortstack{0.93\\{[0.91, 0.95]}} \\

\midrule

\textbf{OUD}
& MaxViT-Tiny
& \shortstack{95.12\\{[87.80, 100.00]}}
& \shortstack{100.00\\{[100.00, 100.00]}}
& \shortstack{94.97\\{[86.80, 100.00]}}
& \shortstack{95.12\\{[87.80, 100.00]}}
& \shortstack{0.99\\{[0.96, 1.00]}} \\

& EfficientNet-B7
& \shortstack{90.24\\{[80.49, 97.56]}}
& \shortstack{100.00\\{[100.00, 100.00]}}
& \shortstack{87.54\\{[78.05, 97.56]}}
& \shortstack{90.24\\{[80.49, 97.56]}}
& \shortstack{0.99\\{[0.96, 1.00]}} \\

& Swin
& \shortstack{87.80\\{[78.05, 97.56]}}
& \shortstack{100.00\\{[100.00, 100.00]}}
& \shortstack{85.63\\{[73.17, 95.12]}}
& \shortstack{87.80\\{[78.05, 97.56]}}
& \shortstack{0.96\\{[0.89, 1.00]}} \\

& ResNet18
& \shortstack{85.37\\{[73.17, 95.12]}}
& \shortstack{100.00\\{[100.00, 100.00]}}
& \shortstack{81.30\\{[71.16, 93.23]}}
& \shortstack{85.37\\{[73.17, 95.12]}}
& \shortstack{0.87\\{[0.75, 0.97]}} \\

\bottomrule
\end{tabular}

\end{table}

\subsection{Lesion-guided ROI-based deep learning performance}\label{subsec3.2}

Lesion-guided ROI-centered crops with controlled perilesional padding produced stronger discrimination performance compared with the global image-based DL framework as shown in Table~\ref{tab:tab2}. On MMOTU, top-1 accuracy ranged from 85.7\% to 93.1\%, top-2 accuracy from 94.0\% to 98.3\%, and AUC from 0.98 to 0.99, with MaxViT-Tiny achieving the highest top-1 accuracy of 93.1\%.

On OUD, top-1 accuracy ranged from 92.7\% to 97.6\%, top-2 accuracy was uniformly 100\%, and AUC ranged from 0.97 to 0.99, with MaxViT-Tiny attaining the highest top-1 accuracy of 97.6\%.

Clinically, this improvement over global image-based analysis demonstrates that directing model attention toward the lesion while preserving limited adjacent structures, such as the cyst wall, septa, papillary projections, and solid components, aligns closely with the lesion characteristics routinely assessed using IOTA and O-RADS descriptors. This lesion-focused strategy substantially reduced classification errors among visually similar entities and may support more confident and objective classification, particularly for less experienced operators.

\begin{table}[htbp]
\caption{Performance of the lesion-guided ROI-based deep learning models on the MMOTU and OUD datasets. Values in brackets represent 95\% confidence intervals.}
\label{tab:tab2}%

\centering
\fontsize{6.5}{7.5}\selectfont
\setlength{\tabcolsep}{2pt}
\renewcommand{\arraystretch}{1.15}

\begin{tabular}{@{}llccccc@{}}
\toprule

\textbf{Data} &
\textbf{Model} &
\shortstack{\textbf{Top-1}\\\textbf{Accuracy (\%)}} &
\shortstack{\textbf{Top-2}\\\textbf{Accuracy (\%)}} &
\shortstack{\textbf{Specificity}\\\textbf{(\%)}} &
\shortstack{\textbf{Sensitivity}\\\textbf{(\%)}} &
\textbf{AUC} \\

\midrule

\textbf{MMOTU}
& MaxViT-Tiny
& \shortstack{93.10\\{[89.66, 96.12]}}
& \shortstack{96.98\\{[94.83, 99.14]}}
& \shortstack{98.99\\{[94.66, 100.00]}}
& \shortstack{93.10\\{[89.66, 96.12]}}
& \shortstack{0.99\\{[0.98, 0.99]}} \\

& EfficientNet-B7
& \shortstack{88.79\\{[84.48, 92.67]}}
& \shortstack{96.55\\{[93.97, 98.71]}}
& \shortstack{98.91\\{[94.51, 100.00]}}
& \shortstack{88.79\\{[84.48, 92.67]}}
& \shortstack{0.99\\{[0.98, 0.99]}} \\

& Swin
& \shortstack{87.07\\{[82.76, 90.95]}}
& \shortstack{98.28\\{[96.55, 99.57]}}
& \shortstack{97.77\\{[92.86, 100.00]}}
& \shortstack{87.07\\{[82.76, 90.95]}}
& \shortstack{0.99\\{[0.98, 0.99]}} \\

& ResNet18
& \shortstack{85.78\\{[81.03, 90.10]}}
& \shortstack{93.97\\{[90.95, 96.56]}}
& \shortstack{98.49\\{[93.66, 100.00]}}
& \shortstack{85.78\\{[81.03, 90.10]}}
& \shortstack{0.97\\{[0.96, 0.98]}} \\

\midrule

\textbf{OUD}
& MaxViT-Tiny
& \shortstack{97.56\\{[92.68, 100.00]}}
& \shortstack{100.00\\{[100.00, 100.00]}}
& \shortstack{98.09\\{[93.58, 100.00]}}
& \shortstack{97.56\\{[92.68, 100.00]}}
& \shortstack{0.99\\{[0.98, 1.00]}} \\

& EfficientNet-B7
& \shortstack{92.68\\{[82.93, 100.00]}}
& \shortstack{100.00\\{[100.00, 100.00]}}
& \shortstack{94.28\\{[90.81, 100.00]}}
& \shortstack{92.68\\{[82.93, 100.00]}}
& \shortstack{0.97\\{[0.92, 1.00]}} \\

& Swin
& \shortstack{95.12\\{[87.80, 100.00]}}
& \shortstack{100.00\\{[100.00, 100.00]}}
& \shortstack{96.18\\{[92.58, 100.00]}}
& \shortstack{95.12\\{[87.80, 100.00]}}
& \shortstack{0.97\\{[0.92, 1.00]}} \\

& ResNet18
& \shortstack{92.68\\{[82.93, 100.00]}}
& \shortstack{100.00\\{[100.00, 100.00]}}
& \shortstack{94.18\\{[90.31, 100.00]}}
& \shortstack{92.68\\{[82.93, 100.00]}}
& \shortstack{0.99\\{[0.97, 1.00]}} \\

\bottomrule
\end{tabular}

\end{table}

\subsection{Lesion contour-based deep learning performance}\label{subsec3.3}

Strict contour-based isolation of the lesions yielded robust performance comparable to that of the lesion-guided ROI-based DL framework as shown in Table~\ref{tab:tab3}. On MMOTU, top-1 accuracy ranged from 85.8\% to 91.8\%, top-2 accuracy from 94.0\% to 97.0\%, and AUC from 0.98 to 0.99, with MaxViT-Tiny achieving the highest top-1 accuracy of 91.8\%.

On OUD, top-1 accuracy ranged from 92.7\% to 95.1\%, top-2 accuracy reached 100\%, and AUC ranged from 0.98 to 0.99. Although performance remained high, the absence of consistent gains over the simpler lesion-guided ROI-based approach indicates that complete removal of the perilesional context may discard subtle supportive features.

This finding has practical relevance because pixel-level lesion contouring is time-intensive and may be difficult to scale in high-volume clinical settings, whereas lesion-guided ROI extraction provides comparable classification performance with a markedly lower annotation burden.

\begin{table}[htbp]
\caption{Performance of the lesion contour-based deep learning models on the MMOTU and OUD datasets. Values in brackets represent 95\% confidence intervals.}
\label{tab:tab3}%

\centering
\fontsize{6.5}{7.5}\selectfont
\setlength{\tabcolsep}{2pt}
\renewcommand{\arraystretch}{1.15}

\begin{tabular}{@{}llccccc@{}}
\toprule

\textbf{Dataset} &
\textbf{Model} &
\shortstack{\textbf{Top-1}\\\textbf{Accuracy (\%)}} &
\shortstack{\textbf{Top-2}\\\textbf{Accuracy (\%)}} &
\shortstack{\textbf{Specificity}\\\textbf{(\%)}} &
\shortstack{\textbf{Sensitivity}\\\textbf{(\%)}} &
\textbf{AUC} \\

\midrule

\textbf{MMOTU}
& MaxViT-Tiny
& \shortstack{91.81\\{[88.36, 95.26]}}
& \shortstack{96.98\\{[94.83, 99.14]}}
& \shortstack{98.97\\{[95.31, 100.00]}}
& \shortstack{91.81\\{[88.36, 95.26]}}
& \shortstack{0.99\\{[0.98, 0.99]}} \\

& EfficientNet-B7
& \shortstack{91.38\\{[87.50, 94.40]}}
& \shortstack{95.69\\{[92.67, 98.28]}}
& \shortstack{99.00\\{[96.01, 100.00]}}
& \shortstack{91.38\\{[87.50, 94.40]}}
& \shortstack{0.99\\{[0.9855, 0.99]}} \\

& Swin
& \shortstack{86.21\\{[81.47, 90.52]}}
& \shortstack{96.98\\{[94.83, 98.72]}}
& \shortstack{97.87\\{[94.31, 100.00]}}
& \shortstack{86.21\\{[81.47, 90.52]}}
& \shortstack{0.99\\{[0.98, 0.99]}} \\

& ResNet18
& \shortstack{85.78\\{[81.03, 90.10]}}
& \shortstack{93.97\\{[90.95, 96.56]}}
& \shortstack{98.49\\{[95.10, 100.00]}}
& \shortstack{85.78\\{[81.03, 90.10]}}
& \shortstack{0.98\\{[0.97, 0.99]}} \\

\midrule

\textbf{OUD}
& MaxViT-Tiny
& \shortstack{95.12\\{[87.80, 100.00]}}
& \shortstack{100.00\\{[100.00, 100.00]}}
& \shortstack{96.18\\{[89.10, 100.00]}}
& \shortstack{95.12\\{[87.80, 100.00]}}
& \shortstack{0.99\\{[0.98, 1.00]}} \\

& EfficientNet-B7
& \shortstack{95.12\\{[87.80, 100.00]}}
& \shortstack{100.00\\{[100.00, 100.00]}}
& \shortstack{96.18\\{[89.10, 100.00]}}
& \shortstack{95.12\\{[87.80, 100.00]}}
& \shortstack{0.99\\{[0.97, 1.00]}} \\

& Swin
& \shortstack{95.12\\{[87.80, 100.00]}}
& \shortstack{100.00\\{[100.00, 100.00]}}
& \shortstack{96.18\\{[88.10, 100.00]}}
& \shortstack{95.12\\{[87.80, 100.00]}}
& \shortstack{0.98\\{[0.93, 1.00]}} \\

& ResNet18
& \shortstack{92.68\\{[82.93, 100.00]}}
& \shortstack{100.00\\{[100.00, 100.00]}}
& \shortstack{94.24\\{[86.10, 100.00]}}
& \shortstack{92.68\\{[82.93, 100.00]}}
& \shortstack{0.99\\{[0.97, 1.00]}} \\

\bottomrule
\end{tabular}

\end{table}

\subsection{Lesion contour-based radiomics performance}\label{subsec3.4}

Radiomics models produced moderate classification performance as shown in Table~\ref{tab:tab4}. On MMOTU, top-1 accuracy ranged from 71.55\% to 74.2\%, while top-2 accuracy ranged from 84.0\% to 87.5\%. Among the evaluated machine learning classifiers, the SVM with a radial basis function kernel achieved the highest top-1 accuracy of 74.2\%.

In OUD, an ANOVA-based feature-selection sweep was performed across different numbers of retained features. Based on this analysis, $k=60$ was selected as the final feature set, with SVM, KNN, and ANN each achieving 95.12\% top-1 accuracy as shown in the supplementary file Fig. S1. Using these 60 selected features, top-2 accuracy reached 100\%. These results highlight radiomics as an interpretable, low-resource, complementary approach that may be suitable for binary confirmation tasks or for integration into existing reporting systems. However, the performance gap compared with DL on the multiclass MMOTU task indicates that fixed quantitative descriptors alone may be insufficient to capture the complex and nonlinear imaging patterns required for reliable eight-class differentiation among overlapping ovarian lesion types.

\begin{table}[htbp]
\caption{Performance of the lesion contour-based radiomics classifiers on the MMOTU and OUD datasets. Values in brackets represent 95\% confidence intervals.}
\label{tab:tab4}%

\centering
\fontsize{6.5}{7.5}\selectfont
\setlength{\tabcolsep}{2pt}
\renewcommand{\arraystretch}{1.15}

\begin{tabular}{@{}llccccc@{}}
\toprule

\textbf{Dataset} &
\textbf{Classifier} &
\shortstack{\textbf{Top-1}\\\textbf{Accuracy (\%)}} &
\shortstack{\textbf{Top-2}\\\textbf{Accuracy (\%)}} &
\shortstack{\textbf{Specificity}\\\textbf{(\%)}} &
\shortstack{\textbf{Sensitivity}\\\textbf{(\%)}} &
\textbf{AUC} \\

\midrule

\textbf{MMOTU}
& ANN
& \shortstack{72.41\\{[66.81, 77.59]}}
& \shortstack{84.05\\{[79.31, 88.36]}}
& \shortstack{94.83\\{[93.51, 96.08]}}
& \shortstack{72.41\\{[66.81, 77.59]}}
& \shortstack{0.92\\{[0.90, 0.94]}} \\

& KNN
& \shortstack{71.55\\{[65.09, 76.72]}}
& \shortstack{84.91\\{[80.17, 89.23]}}
& \shortstack{94.32\\{[92.78, 95.60]}}
& \shortstack{71.55\\{[65.09, 76.72]}}
& \shortstack{0.91\\{[0.88, 0.93]}} \\

& SVM-RBF
& \shortstack{74.28\\{[67.67, 78.88]}}
& \shortstack{87.50\\{[83.19, 91.38]}}
& \shortstack{94.93\\{[93.80, 96.10]}}
& \shortstack{74.28\\{[67.67, 78.88]}}
& \shortstack{0.93\\{[0.91, 0.95]}} \\

\midrule

\textbf{OUD}
& ANN
& \shortstack{95.12\\{[87.80, 100.00]}}
& \shortstack{100.00\\{[100.00, 100.00]}}
& \shortstack{96.18\\{[89.45, 100.00]}}
& \shortstack{95.12\\{[87.80, 100.00]}}
& \shortstack{0.98\\{[0.93, 1.00]}} \\

& KNN
& \shortstack{95.12\\{[87.80, 100.00]}}
& \shortstack{100.00\\{[100.00, 100.00]}}
& \shortstack{96.18\\{[89.45, 100.00]}}
& \shortstack{95.12\\{[87.80, 100.00]}}
& \shortstack{0.97\\{[0.92, 1.00]}} \\

& SVM-RBF
& \shortstack{95.12\\{[87.80, 100.00]}}
& \shortstack{100.00\\{[100.00, 100.00]}}
& \shortstack{96.18\\{[89.45, 100.00]}}
& \shortstack{95.12\\{[87.80, 100.00]}}
& \shortstack{0.95\\{[0.86, 1.00]}} \\

\bottomrule
\end{tabular}

\end{table}

\subsection{Overall comparison of classification strategies}\label{subsec3.5}

Across all evaluated frameworks, the lesion-guided ROI-based and lesion contour-based DL strategies achieved the strongest overall performance. MaxViT-Tiny delivered the highest top-1 accuracies, reaching 93.1\% on MMOTU and 97.6\% on OUD, with corresponding AUC values exceeding 0.99, confirming excellent threshold-independent discrimination.

\begin{figure}[htbp]
\centering
\includegraphics[width=\textwidth]{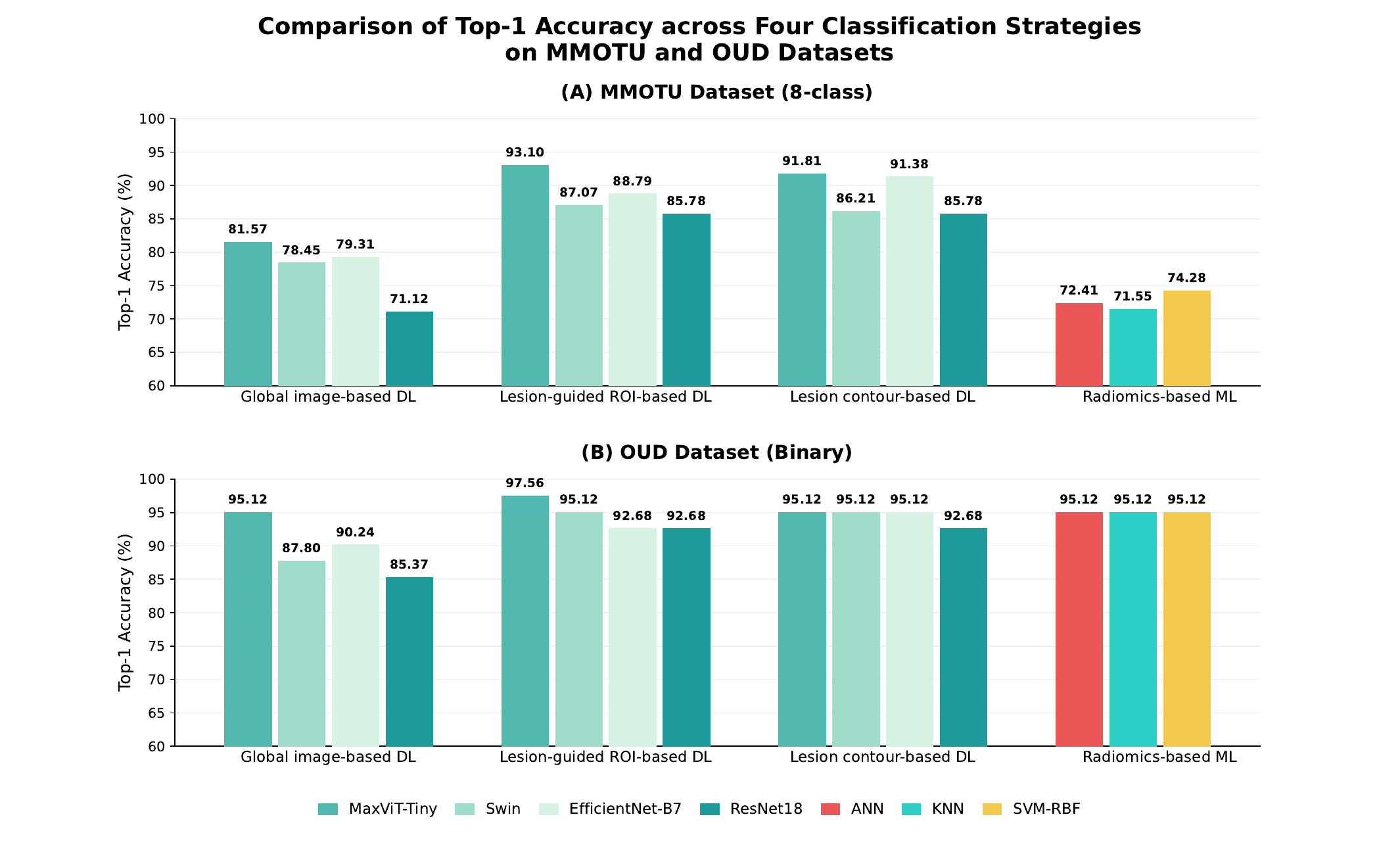}
\caption{Comparison of top-1 accuracies across all evaluated models and classification strategies. }
\label{fig4}
\end{figure}

Compared with the global image-based framework, lesion-focused strategies improved classification performance, particularly on the multiclass MMOTU dataset. The lesion-guided ROI approach achieved the highest overall top-1 accuracy, while contour-based DL showed comparable performance but required more detailed annotation. As summarized in Fig.~\ref{fig4}, lesion-guided ROI cropping provided the most practical balance between diagnostic performance and annotation burden.

\section{Discussion}\label{sec4}

 This study compared four ovarian ultrasound classification strategies across two open-access datasets representing multiclass and binary classification tasks. Overall, the lesion-guided ROI-based strategy provided the most favorable balance between classification performance and annotation burden. Because TVS assessment relies substantially on morphological characteristics within the lesion–tissue interface, a modest perilesional margin retains these cues while excluding distant background and acquisition artifacts.

This finding has potential clinical relevance for reducing operator dependency, which remains an important source of inter-observer variability in ultrasound interpretation. The gain from global to lesion-guided inputs indicates that explicit ROI guidance improves capture of textural and structural distinctions among lesions with overlapping appearances, which may yield more consistent outputs and support less experienced operators in triage, follow-up, and referral.

\begin{figure}[htbp]
\centering
\includegraphics[width=\textwidth]{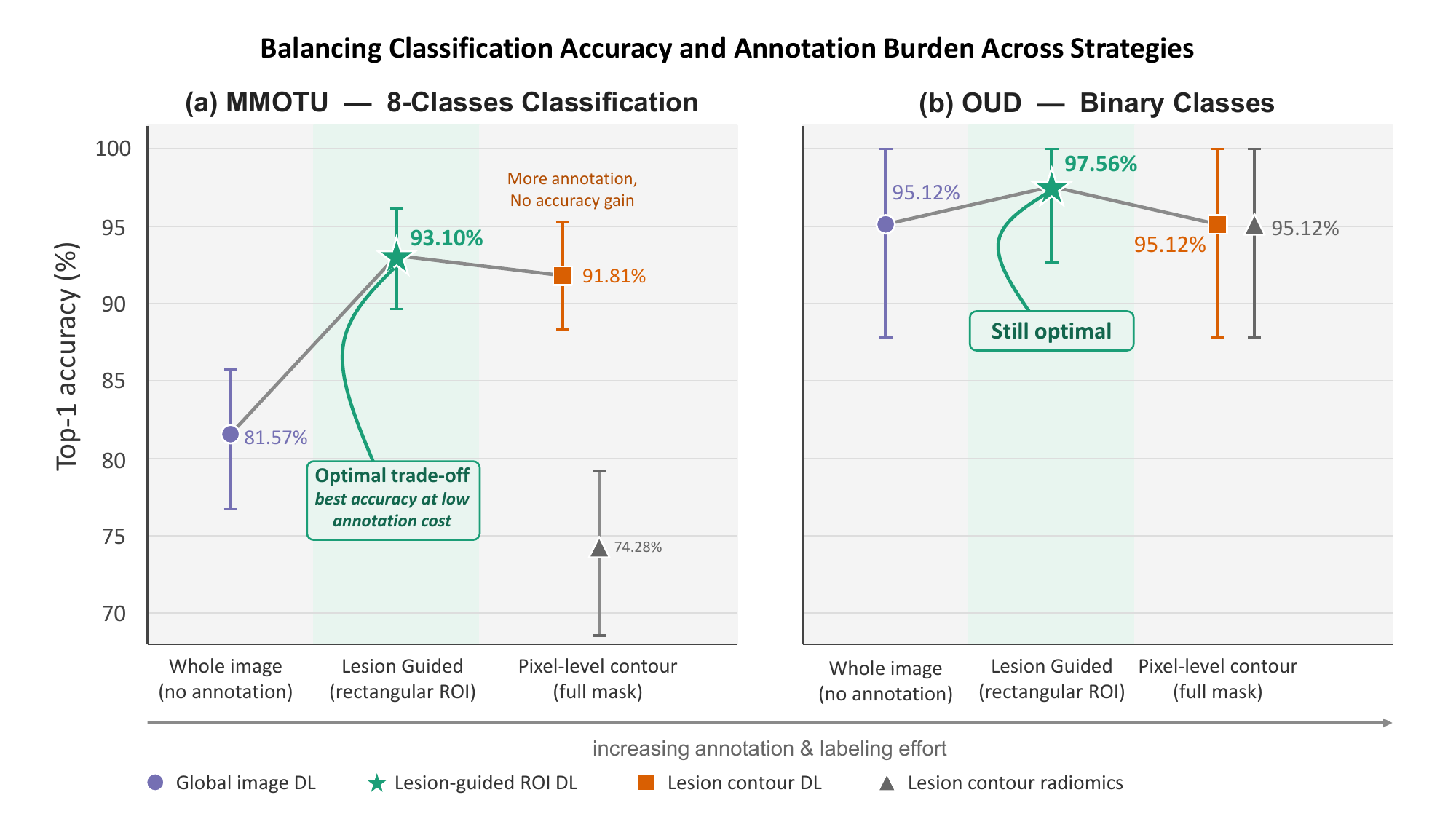}
\caption{Effect of lesion localization strategy and annotation complexity on ovarian ultrasound classification performance.}
\label{fig5}
\end{figure}

 From a practical perspective, improved classification performance did not require highly detailed pixel-level lesion contouring. The lesion-guided ROI-based DL framework achieved the strongest overall performance using a simpler lesion-centered crop that preserved relevant tumor information and limited surrounding context while reducing annotation burden, as illustrated in Fig.~\ref{fig5}. This makes lesion-guided ROI annotation more practical for large-scale dataset development and clinical workflows than precise contour annotation. Global image-based DL may be affected by irrelevant image regions, whereas radiomics offers a resource-efficient and interpretable complement to DL. However, its lower multiclass performance suggests that handcrafted and quantitative features may not fully capture the complex patterns needed for fine-grained ovarian lesion classification.

 Previous studies using the MMOTU and OUD datasets have demonstrated the potential of AI for ovarian ultrasound classification; however, systematic comparisons of different lesion representation strategies remain limited, as shown in Table~\ref{tab:comparison}. On MMOTU, \cite{bib23} reached 80.60\% top-1 accuracy, \cite{bib24} reached 91.9\% accuracy, and \cite{bib25} reached 70.6\% mAP50, whereas our lesion-guided MaxViT-Tiny achieved 93.10\% accuracy on the same eight-class task.
 
 The original OUD study \cite{bib26} reported 76.2\% accuracy using ResNet18 for three-class classification. However, our study reformulated OUD as a binary DF-versus-PCO task because of the limited number of normal-ovary cases, so the two results are not directly comparable. To enable a fair within-study comparison, ResNet18 was retrained under the same binary setting, achieving 85.37\% accuracy with global images and 92.68\% with lesion-guided ROI inputs, corresponding to a 7.31\% improvement with lesion guidance. Notably, contour-based ResNet18 achieved the same 92.68\% accuracy, indicating that precise contouring did not provide an additional top-1 accuracy gain. Furthermore, lesion-guided MaxViT-Tiny achieved the highest accuracy of 97.56\%, outperforming lesion-guided ResNet18 by 4.88\%

\begin{table}[htbp]
\caption{Comparison of the proposed approach with previous studies on the MMOTU and OUD datasets.}
\label{tab:comparison}%

\centering
\fontsize{6.5}{7.5}\selectfont
\setlength{\tabcolsep}{2pt}
\renewcommand{\arraystretch}{1.15}

\begin{tabular}{@{}llllcc@{}}
\toprule

\shortstack{\textbf{Study}\\\textbf{(Year)}} &
\textbf{Dataset} &
\textbf{Task} &
\textbf{Method} &
\shortstack{\textbf{Best}\\\textbf{Accuracy}} &
\textbf{AUC} \\

\midrule

(2024) \cite{bib25}
& \shortstack{MMOTU\\(public)}
& 8-class
& \parbox[t]{4.0cm}{\raggedright YOLOv8x\par}
& \shortstack{70.6\%\\(mAP50)}
& -- \\

(2025) \cite{bib23}
& \shortstack{MMOTU\\(public)}
& 8-class
& \parbox[t]{4.0cm}{\raggedright EfficientNetV2-M (whole image)\par}
& 80.60\%
& -- \\

(2026) \cite{bib24}
& \shortstack{MMOTU\\(public)}
& 8-class
& \parbox[t]{4.0cm}{\raggedright EfficientOvaNet (EfficientNet-B3, ROI + global; pixel masks)\par}
& 91.9\%
& 0.980 \\

\textbf{Proposed}
& \shortstack{\textbf{MMOTU}\\\textbf{(public)}}
& \textbf{8-class}
& \parbox[t]{4.0cm}{\raggedright\textbf{Lesion-guided ROI + MaxViT-Tiny (single model, bounding box)}\par}
& \textbf{93.10\%}
& \textbf{0.991} \\

\midrule

(2025) \cite{bib26}
& \shortstack{OUD\\(public)}
& \shortstack{3-class\\(normal/DF/PCO)}
& \parbox[t]{4.0cm}{\raggedright ResNet18\par}
& 76.2\%
& -- \\

(2025) \cite{bib26}
& \shortstack{OUD\\(Binary)}
& \shortstack{2-class\\DF vs PCO}
& \parbox[t]{4.0cm}{\raggedright Global image ResNet18 architecture-matched baseline\par}
& 85.37\%
& 0.879 \\

\textbf{Proposed}
& \shortstack{OUD\\(Binary)}
& \shortstack{2-class\\DF vs PCO}
& \parbox[t]{4.0cm}{\raggedright Lesion-guided ROI ResNet18\par}
& 92.68\%
& 0.995 \\

\textbf{Proposed}
& \shortstack{OUD\\(Binary)}
& \shortstack{2-class\\DF vs PCO}
& \parbox[t]{4.0cm}{\raggedright Lesion contour ResNet18\par}
& 92.68\%
& 0.995 \\

\textbf{Proposed}
& \shortstack{\textbf{OUD}\\\textbf{(public)}}
& \shortstack{\textbf{2-class}\\\textbf{DF vs PCO}}
& \parbox[t]{4.0cm}{\raggedright\textbf{Lesion-guided ROI + MaxViT-Tiny}\par}
& \textbf{97.56\%}
& \textbf{0.997} \\

\bottomrule
\end{tabular}

\end{table}

Together, these comparisons suggest that the strategy used to define the learning region is an important determinant of classification performance. A simple lesion-guided ROI achieved performance comparable to, and in some settings better than, precise contour-based isolation while requiring less detailed annotation. The systematic comparison of all four strategy under a consistent experimental protocol therefore highlights lesion representation as an important consideration in ovarian ultrasound AI development and supports lesion-guided ROI localization as a potentially scalable approach.

This study has several limitations. The analysis was based on public datasets and B-mode ultrasound images; therefore, further validation using heterogeneous clinical data, additional imaging modalities such as Doppler or contrast-enhanced ultrasound, and evaluation of computational requirements are needed before clinical implementation.

Future studies should investigate annotation-efficient approaches, including weakly supervised learning and automated lesion localization, as well as multimodal integration of ultrasound imaging with relevant clinical information. Prospective multicenter validation across different clinical settings, patient populations, and ultrasound systems will be essential to determine the generalizability and real-world applicability of the proposed framework.

\section{Conclusion}\label{sec5}

This study suggests that a lesion-guided ROI DL strategy may offer a practical balance between diagnostic performance and annotation feasibility for ovarian ultrasound classification. By directing attention to the lesion while retaining limited surrounding context, this approach could support more consistent interpretation with reduced clinician burden compared with global or strictly contour-based methods. Prospective multicenter validation is needed to confirm the performance of the proposed technique and its potential clinical utility.

\backmatter

\bmhead{Supplementary information}

Supplementary material is available with the online version of this article.

\section*{Declarations}

\subsection*{Funding}
This research has been funded by FTI-Dissertationen grant with project number: FTI24-D-008. 

\subsection*{Competing interests}
The authors declare no competing interests.

\subsection*{Ethics approval}
This study involved secondary analysis of publicly available, de-identified ovarian ultrasound datasets. No new participants were recruited, and no new patient data were collected for this study. Ethical approval and patient consent for the original data collection were reported by the respective dataset providers.

\subsection*{Data availability}
The datasets analyzed in this study are publicly available. The Multi-Modality Ovarian Tumor Ultrasound (MMOTU) dataset is described in \cite{bib23}, and the Ovarian Ultrasound Dataset (OUD) is described in \cite{bib26}.






\bibliography{sn-bibliography}


\end{document}